\pdfoutput=1

\PassOptionsToPackage{sort}{natbib}

\PassOptionsToPackage{inline}{enumitem}

\documentclass[11pt]{article}

\usepackage[]{acl}

\usepackage{fancyhdr}
\usepackage{times}
\usepackage{latexsym}

\usepackage[T1]{fontenc}

\usepackage[utf8]{inputenc}

\usepackage{microtype}

\usepackage{enumitem}
\usepackage[para]{threeparttable}
\usepackage{booktabs}
\usepackage{multirow}
\usepackage{graphicx}
\usepackage{amsmath}
\usepackage{amssymb}
\usepackage{hyperref}
\usepackage{xcolor}
\usepackage{colortbl}
\usepackage{longtable}
\usepackage{circledsteps}
\usepackage{longfbox} 
\usepackage{bm}

\definecolor{patterncolor}{HTML}{e6f2e6}
\definecolor{patterncolor2}{HTML}{d5eef2}
\definecolor{patterncolor3}{HTML}{f5e9d0}

\definecolor{patterncolor4}{HTML}{ccdde0}
\definecolor{patterncolor5}{HTML}{e2d5e8}

\definecolor{circlenumbercolor}{HTML}{B83676}

\makeatletter
\newdimen\@tempdimd
\makeatother

\newfboxstyle{patternparam}{padding=2pt, margin-bottom=1pt, margin-top=1pt, border-radius=3pt, border-style=none, height=9pt, background-color=patterncolor2}

\newfboxstyle{boxinsitu}{padding=2pt, padding-bottom=1pt, margin-bottom=1pt, margin-top=1pt, border-radius=3pt, border-style=none, height=9pt, background-color=patterncolor4}

\newfboxstyle{boxdaily}{padding=2pt, padding-bottom=1pt, margin-bottom=1pt, margin-top=1pt, border-radius=3pt, border-style=none, height=9pt, background-color=patterncolor5}

\newcommand\myCircled[2][]{
\Circled[inner ysep=5pt, fill color=circlenumbercolor, outer color=circlenumbercolor, inner color=white,#1]{\scriptsize\sffamily\textbf{#2}}
}

\newcommand\circledBody[2][]{
\hspace{-3pt}\Circled[inner ysep=5pt, fill color=circlenumbercolor, outer color=circlenumbercolor, inner color=white,#1]{\small\sffamily\textbf{#2}}\hspace{-3pt}
}

\definecolor{revisedcolor}{RGB}{0,0,255}

\definecolor{tableheader}{HTML}{EFEFEF}

\definecolor{tabledarkheader}{RGB}{128,128,128}
\definecolor{tablegrayline}{HTML}{d0d0d0}

\newcommand{\thicktablehline}{\arrayrulecolor{black}\specialrule{0.75pt}{2pt}{0pt}}

\newcommand{\tablelighthline}{\arrayrulecolor{tablegrayline}\hline}

\definecolor{trackerrowbackground}{HTML}{f9f9f9}
\newcommand{\trackerrowcolor}{\rowcolor{trackerrowbackground}}

\renewcommand{\thefootnote}{\alph{footnote}}

\newcommand{\astfootnote}[1]{%
\let\oldthefootnote=\thefootnote%
\setcounter{footnote}{0}%
\renewcommand{\thefootnote}{\fnsymbol{footnote}}%
\footnote{#1}%
\let\thefootnote=\oldthefootnote%
}

\definecolor{revisedcolor}{RGB}{0,0,0}

\newcommand{\revised}[1]{\textcolor{revisedcolor}{#1}}

\title{Leveraging Pre-Trained Language Models to Streamline \protect\\ Natural Language Interaction for Self-Tracking}

\author{Young-Ho Kim$^\P$ \hspace{7mm} Sungdong Kim$^\P$ \hspace{7mm} $^*$Minsuk Chang$^\P$ \hspace{7mm} Sang-Woo Lee$^\P$$^\ddagger$ \\
  $^\P$NAVER AI Lab, $^\ddagger$NAVER CLOVA\\
  \{ygho.kim, sungdong.kim, minsuk.chang, sang.woo.lee\}@navercorp.com\\}
  
\newcommand{\fone}{$F_{1}$}

\newcommand{\eg}{\textit{e.g.}}
\newcommand{\ie}{\textit{i.e.}}
\newcommand{\cf}{\textit{c.f.}}

\begin{document}

\twocolumn[{
\renewcommand\twocolumn[1][]{#1}
\maketitle
\begin{center}
    \vspace{-1.2 cm}
    \centering
    \includegraphics[width=1.0\textwidth]{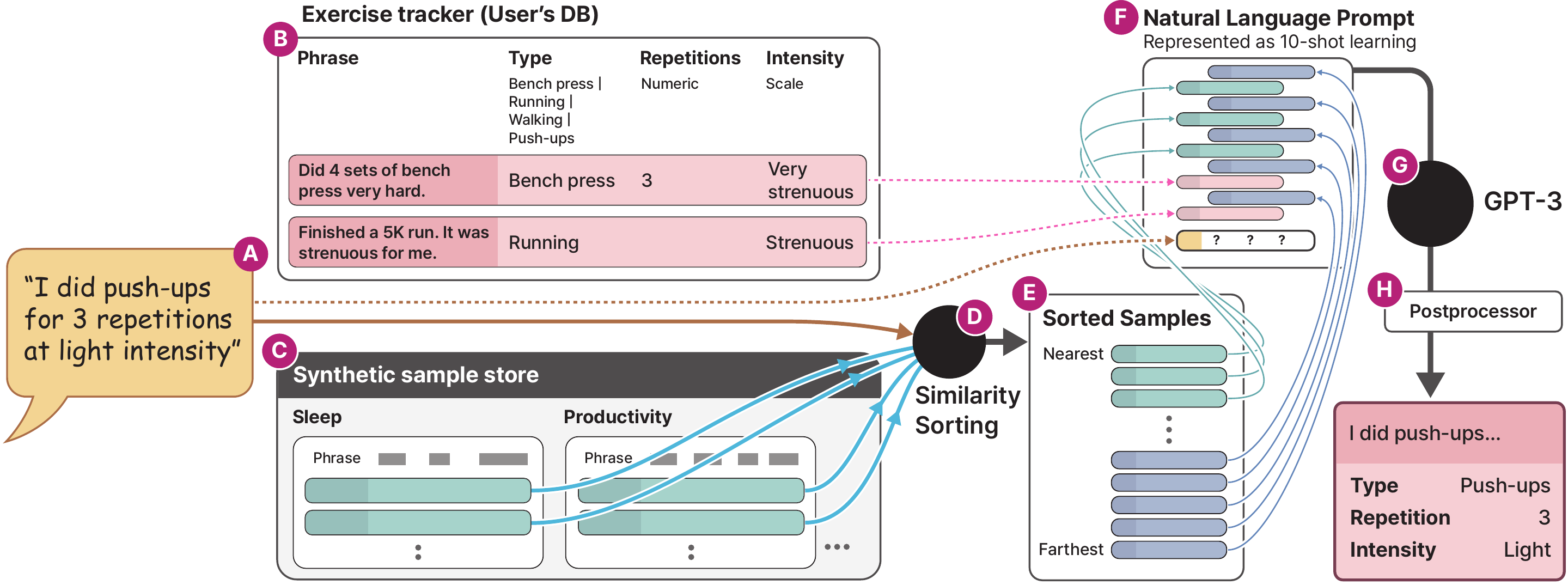}
    \captionof{figure}{A two-shot learning example on our NLU framework, when the person has two prior items for exercise tracker~\myCircled{B}. The system augments the prompt with synthetic seed samples~\myCircled{C} to transform the task into a 10-shot learning problem~\myCircled{F}. Refer to \autoref{appendix:prompt} for an example prompt.}
    
    \vspace{0.3 cm}
    \label{fig:framework}
\end{center}
}]

\thispagestyle{fancy}

\footnotetext{\hspace{-2mm}$^*$Minsuk is now at Google.}
\setcounter{footnote}{0}
\renewcommand{\thefootnote}{\arabic{footnote}}

\begin{abstract}
Current natural language interaction for self-tracking tools largely depends on \revised{bespoke implementation optimized for a specific tracking theme and data format}, which is neither generalizable nor scalable to a tremendous design space of self-tracking. However, training machine learning models in the context of self-tracking is challenging due to the wide variety of tracking topics and data formats.
In this paper, we propose a novel NLP task for self-tracking that extracts close- and open-ended information from a retrospective activity log described as a plain text, and a domain-agnostic, GPT-3-based NLU framework that performs this task. The framework augments the prompt using synthetic samples to transform the task into 10-shot learning, to address a cold-start problem in bootstrapping a new tracking topic. Our preliminary evaluation suggests that our approach significantly outperforms the baseline QA models. Going further, we discuss future application domains toward which the NLP and HCI researchers can collaborate.
\end{abstract}

\section{Introduction}
Self-tracking tools (\eg, mHealth apps like Fitbit~[\citealt{Fitbit}]) help people longitudinally track their health and activity in a structured and systematic manner.
The advancement of Natural Language Interaction (NLI) techniques have opened new opportunities for designing novel self-tracking systems with which people can intuitively record their data using speech and/or chat; specifying long and complex information in natural language is generally more flexible and expressive than using predetermined forms in traditional graphical widgets~\cite{Luo2021FoodScrap, Kim2021Data@Hand, Luo2020TandemTrack}. As a result, there is a growing interest in building \textit{speech-mediated self-tracking} tools to offer low-burden~(\eg,~\citealt{Luo2021FoodScrap}) and accessible~(\eg,~\citealt{Kim2022MyMove}) self-tracking.

\revised{
Yet, existing systems predominantly incorporate bespoke (and mostly rule-based) natural language understanding (NLU) logics optimized for capturing uniform information in a specific tracking theme, compromising generalizability and scalability to diverse user inputs and contexts.}
For example, the NLU of Data@Hand, a visual exploration app for fitness data, is implemented using syntax-based rules with POS (part of speech) tags and predefined keywords~\cite{Kim2021Data@Hand}. Such an approach is known to be vulnerable to a selection of vocabulary and individualized linguistic patterns~\cite{Kim2021Data@Hand, Kim2019VoiceCuts}. Furthermore, extending the NLU to support a new type of data (e.g., exercise sessions) \revised{is a demanding task} because it requires appending new rules manually.

\revised{
Despite the promise of deep-learning-based NLP approaches, it is still challenging to develop flexible and scalable NLU models for self-tracking mainly because} 
the design space of self-tracking is broad in terms of topics and data formats~\cite{Kim2017OmniTrack, Epstein2020Mapping}. Hence, it is overwhelming to collect natural language datasets that cover the entire space and varied domains.


In this work, we introduce a novel NLP task for self-tracking focusing on in-situ data collection scenarios where people capture their retrospective activity logs. We also propose a novel NLU framework (\autoref{fig:framework}) that supports this task, which incorporates GPT-3~\cite{Brown2020FewShotLearners}, a large-scale pre-trained language model (PLM), to handle linguistic variations of natural language commands (\eg, ``\textit{I drank a cup of coffee an hour ago}'' or ``\textit{At 3:00 PM, had an Americano.}''). To bootstrap in-context learning on GPT-3, the framework leverages synthetic samples constructed from simulations with 24 seed tracking schemas. Given a natural language phrase, the framework extracts values for data fields from a data table. The phrase may be terse and specify only a subset of data fields.

Our preliminary evaluation shows that our prompt augmentation approach using synthetic seed samples was effective in extracting appropriate information from the input phrases in a low-resource scenario. In pure zero-shot cases, GPT-3 underperformed the T5-based model~\cite{raffel2019exploring, lin2021zero}. However, it outperformed when augmented with synthetic seed samples that had different data schemas and tracking topics. The performance increased by the number of prior examples in the corresponding data schema but saturated with more examples. Our findings demonstrate the opportunities of GPT-3's in-context learning abilities for avoiding a cold start problem of the natural-language-based data collection task.

\section{Background}
Self-tracking is a powerful means of understanding oneself and self-promoting positive behavior changes~\cite{Choe2014QSelferPractice, Li10StageBasedModel}. People capture their activities in a variety of topics including but not limited to physical/mental health, finance, productivity, diet, and sleep~\cite{Epstein2020Mapping}. The data points collected for self-tracking usually describe a phenomenon during a time interval or for an associated time point at a specific granularity like minute or day~\cite{Kim2019HealthDataAccessibility}. The phenomenon information consists of various types of data fields such as numbers (\eg, step count, heart rate), texts (\eg, description of a stress episode), scales (\eg, productivity, stress level, sleep quality) or choices (\eg, type of mood)~\cite{Kim2017OmniTrack, Jeon2016MsThesis}. 

While fitness trackers can capture various health metrics, many of the human activities cannot be captured by sensors and still require a manual input to be captured. 
As an effective way to reduce the manual input burden, the speech modality has recently gained interest and was applied to smart speakers (\eg, Fitbit Skill~[\citealt{FitbitSkill}], MyFitnessPal Skill~[\citealt{MyFitnessPalSkill}]) and research prototypes~(\eg, ModEat~[\citealt{Silva2021Food}], TandemTrack~[\citealt{Luo2020TandemTrack}], FoodScrap~[\citealt{Luo2021FoodScrap}]). These tools support speech-based data capture through smart speakers, smartwatches, or smartphones. 
Our work expands this growing body of speech-based self-tracking research by proposing a unified framework of NLU to support flexible phrasing of multifaceted information in arbitrary tracking topics, which are not yet supported by prior systems. 
\section{NLU Framework for Self-Tracking}


\subsection{Task Description}
Imagine a person uses a self-tracking platform that consists of multiple data tables for sub-topics. This is analogous to common health platforms such as Fitbit~\cite{Fitbit} and Apple Health~\cite{AppleHealth} with multiple data tables for step count, body weights, food, or water intake. We refer to the individual data tables as \textbf{trackers}~(\eg, \circledBody{B} in \autoref{fig:framework}), and data points for each tracker as \textbf{items}~(\eg, colored bars in \circledBody{B} in \autoref{fig:framework}). A tracker comprises multiple input fields with six data types---\textit{number}, \textit{Likert scale}, \textit{single-choice}, \textit{multiple-choice}, \textit{short-form text}, and \textit{long-form text}, which are derived from prominent data types of existing self-tracking apps~\cite{Jeon2016MsThesis, Kim2017OmniTrack}. The person may use natural language to insert a new item to the database. For example, he or she may speak, ``\textit{I did \lfbox[patternparam]{\textbf{push-ups}} for \lfbox[patternparam, background-color=patterncolor]{\textbf{three}} repetitions at \lfbox[patternparam, background-color=patterncolor3]{\textbf{light}} intensity,}'' to describe an item, \textit{\{Exercise $\rightarrow$ \lfbox[patternparam]{push-ups}, Repetitions $\rightarrow$ \lfbox[patternparam, background-color=patterncolor]{3}, Intensity $\rightarrow$ \lfbox[patternparam, background-color=patterncolor3]{light}\}}, upon finishing the exercise session. Such an interaction may be performed via a smartphone app, chatbots, or voice assistants. This main task of our framework can be represented as $itm'_{trk} = [v_{trk}^{0} ... v_{trk}^{i} ... v_{trk}^{n-1}] = NLU(trk, phr)$, where $NLU$ derives a list of value $v$ for $n$ fields of the tracker $trk$ from the phrase $phr$. Assuming that we have little or no instances for $Itm'_{trk}$, we solve $NLU()$ through few-shot learning with GPT-3. We turned to PLM because it can be switched to a different problem upon a new natural language prompt with a handful of examples~\cite{Liu2021PretrainPromptAndPredict}, whereas traditional machine learning models require a large amount of task-specific datasets.

\subsection{Prompt Augmentation}
In the early stage of using $trk$, the person may not have little or no items for it; the system does not have sample instances with the same data schema to be put in a prompt for few-shot learning, when it receives a new phrase for $trk$. To overcome the instability of accuracy in low-shot cases~\cite{Brown2020FewShotLearners}, we transformed the NLP task into a 10-shot learning problem by augmenting the model prompt (\cf, \autoref{appendix:prompt}) using \textit{synthetic samples}. We constructed a synthetic sample store (\circledBody{C} in \autoref{fig:framework}) with 504 item-phrase pairs from 24 trackers (21 pairs per each tracker). The trackers were manually composed by the authors (see \autoref{appendix:seed_trackers} for an exhaustive list of trackers). We randomly generated item samples and phrases that describe the content using GPT-3. Four authors iteratively inspected the data and corrected wrong matches between the values and the phrase. Each sample contains a \textit{subset} of data fields of the tracker, to simulate the cases when people do not include all field values in a single utterance.~(See the second item in \circledBody{B} in \autoref{fig:framework} that omitted \textit{Repetitions}.)

The current implementation mixes both the nearest five and the farthest five samples in a prompt (\circledBody{D}--\circledBody{F} in \autoref{fig:framework}), inspired by \citealt{Liu2021RetrievalGPT3} and \citealt{Zhao2021CalibrateBeforeUse}. We used cosine similarity between the embeddings calculated using a sentence transformer \texttt{multi-qa-MiniLM-L6-cos-v1} in the \texttt{sentence-transformers}\footnote{https://pypi.org/project/sentence-transformers/} package. When there exist items and phrases for the tracker, they are treated as the nearest samples and placed near the output of the prompt (\circledBody{F} in \autoref{fig:framework}).
The framework passes the prompt to GPT-3 via OpenAI's API\footnote{https://openai.com/api/}. \revised{Specifically, we used \texttt{text-davinci-002}, the most capable \textit{InstructGPT}~\cite{Long2022InstructGPT} model optimized for following human prompts.}
Finally, the postprocessor (\circledBody{H} in \autoref{fig:framework}) parses the plain text output into a data table and matches the choice labels to the nearest ones in a tracker schema using the same transformer used in \circledBody{D}.
\section{Preliminary Evaluation}
To obtain preliminary insights on the feasibility of our approach, we evaluated the task outcomes from a series of scenarios, using the synthetic samples as a validation dataset.

\paragraph{Baselines}
We evaluated \textit{TransferQA} and pure zero-shot \textit{GPT-3} as two baseline models.
\revised{Since the proposed task has not been thoroughly explored in the NLP discipline, we chose TransferQA, one of the best-performing model for \textit{dialog state tracking}, whose task is the most similar to the proposed one.}
TransferQA is a T5-based model pre-trained on various question answering (QA) datasets including extractive and multiple-choice QA~\cite{raffel2019exploring, lin2021zero}. Originally, it was proposed for zero-shot dialogue state tracking and utilizes slot description as a question to extract corresponding value from a given input text. 
Although having descriptions for each data field is not realistic in our case, we manually added the description to each field of the trackers to construct input prompts of TransferQA. For example, ``\textit{extractive question: the number of repetitions or laps of the exercise? context: user: i did push-ups for 3 repetitions at light intensity.}'' is the input representation to get \textit{Repetitions} of exercise from the case in \autoref{fig:framework}. For the choice and Likert scale fields, the options were included the prompt as well. 
For GPT-3, we prompted the model to extract field values from an input phrase by giving only the tracker schema without any examples.

\paragraph{In-context Learning}
We simulated the scenarios where the user provides a phrase when there are zero to four prior items in a database for the corresponding tracker. Since we used synthetic samples as a validation set, we treated one of the sample trackers as a user tracker and excluded the samples for the tracker from the store when augmenting the prompt. We iterated over all 24 trackers and 504 items, for each N-shot iteration (504 $\times$ 5 = 2520).

\paragraph{Evaluation Metrics} 
We employed joint goal accuracy (JGA) and \fone~score, which are usually used for the dialogue state tracking tasks to measure the NLU performance of the models. JGA checks whether all predicted values are exactly matched with the ground truth values whereas \fone~checks partial matches between them. For these measures, we excluded \textit{long-form text} fields, for which we instead measured BLEU-4 and ROUGE-L scores.

\paragraph{Results}
\autoref{table:main_eval} illustrates the evaluation results.
In pure zero-shot cases, TransferQA slightly outperformed for close-ended fields (JGA and \fone) but GPT-3 performed almost twice better in extracting open-ended, long-form text fields (B-4 and R-L).
In in-context learning cases, GPT-3 augmented with synthetic samples surprisingly outperformed both baseline models in both close-ended and open-ended fields. Even when there were no prior items for the corresponding tracker (zero-shot in in-context learning), JGA was improved by 16.3\% and \fone~by 18.4 in GPT-3. The preformance generally increased by the increase of the number of prior items but seemed to be saturated around two- or three-shots.

\paragraph{Limitation}
As a preliminary evaluation, we used the synthetic samples as a validation set. For a more ecologically valid evaluation, we need a human-generated dataset in the future.

\begin{table}[t!]
    \centering
    \small
    \sffamily
	\def\arraystretch{1.5}
    \setlength{\tabcolsep}{0.6em}
    \arrayrulecolor{black}
    \begin{tabular}{|p{0.22\columnwidth}!{\color{gray}\vrule}c!{\color{gray}\vrule}cc!{\color{lightgray}\vrule}cc|}
    \hline
    \rowcolor{tableheader}
        \textbf{Model} & \textbf{N-shot} & \textbf{JGA} & \textbf{\textit{F}$\bm{_{1}}$} & $\bm{^a}$\textbf{B-4} & $\bm{^b}$\textbf{R-L}  \\
        \rowcolor[rgb]{0.5,0.5,0.5}
        \multicolumn{6}{l!{\color{black}\vrule}}{\textcolor{white}{\sffamily\textit{Pure Zero-shot (Baseline)}}} \\
        \hline
        \textbf{TransferQA} & 0 & 27.8 & 53.7 & 13.6 & 28.7  \\
        \textbf{GPT-3} & 0 & 26.2 & 49.9 & 41.2 & 58.5 \\
        \rowcolor[rgb]{0.5,0.5,0.5}
        \multicolumn{6}{!{\color{black}\vrule}l}{\textcolor{white}{\sffamily\textit{In-context Learning (Augmented 10-shot Prompts)}}} \\
        & 0 & 42.5 & 68.3 & 56.3 & 77.1 \\
        & 1 & 51.2 & 73.1 & 57.0 & 78.9 \\
        \textbf{GPT-3} & 2 & \textbf{57.7} & 75.6 & \textbf{58.8} & \textbf{80.5} \\
        & 3 & 56.9 & \textbf{76.6} & 57.3 & 78.3 \\
        & 4 & 56.5 & 76.4 & 57.8 & 78.8 \\
        \hline
    \end{tabular}
    \begin{flushright}
    \sffamily\scriptsize\textbf{$\bm{^a}$: BLEU-4, $\bm{^b}$: ROUGE-L}
    \end{flushright}
    \vspace{-3mm}
    \caption{Zero and few-shot evaluation results on our validation dataset. The N-shot is the number of examples of the corresponding tracker included in a prompt. 
    }
    \label{table:main_eval}
\end{table}
\begin{figure*}[t]
    \centering
    \includegraphics[width=\textwidth]{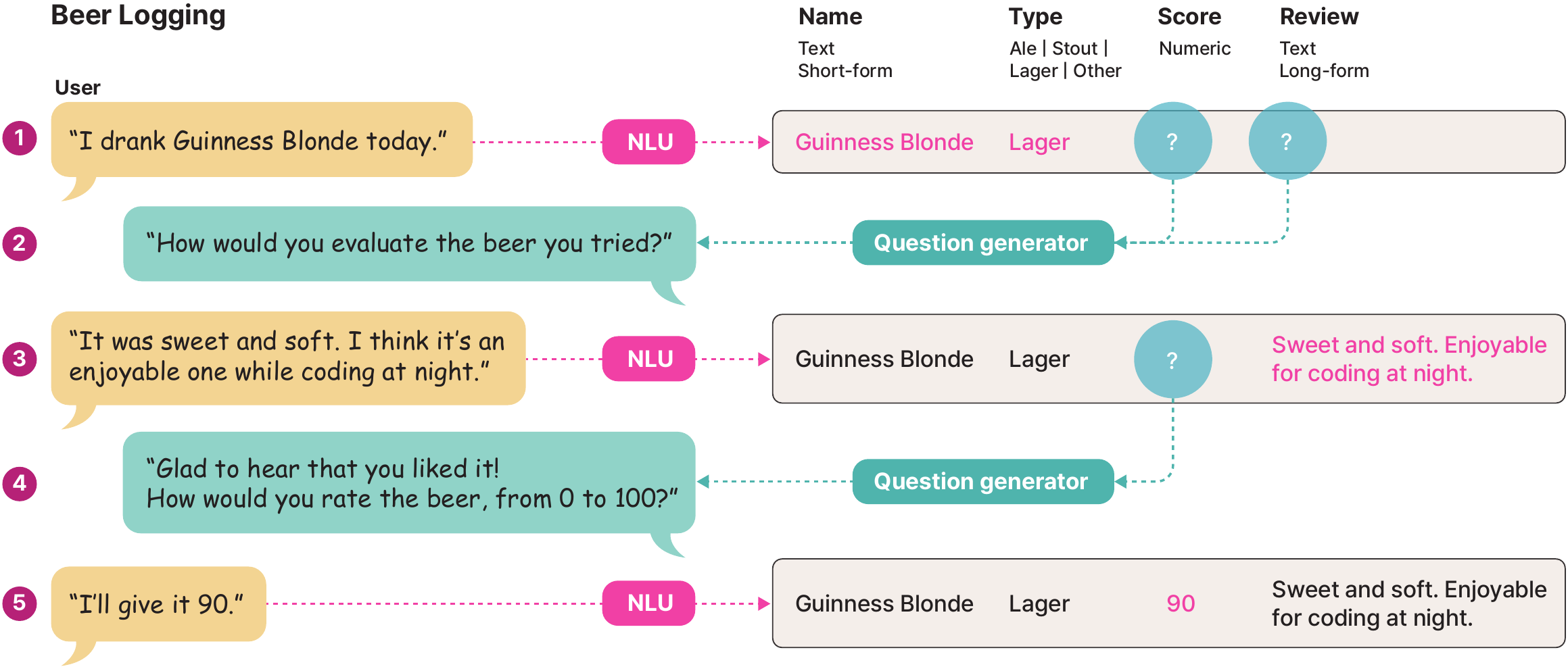}
    \caption{A scenario for a conversational agent embedding our framework (NLU) and a question generator. Upon receiving a user utterance that specifies a subset of field values of the tracker (elements in pink on the right), the question generator suggests a proper response of the system that asks back the user to specify the remaining fields.}
    \label{fig:multiturn}
\end{figure*}

\section{Discussion and Future Directions}
In this section, we discuss the rooms for improvement and envision collaborative application domains for both HCI and NLP researchers.

\subsection{Prompt Engineering and Seed Sampling}
In this work, we mixed the nearest and farthest samples in terms of linguistic similarity between the phrases. A logical next step would be to investigate different strategies to generate prompts. For example, we may hierarchically pick the appropriate trackers first and then retrieve samples from them. Another approach is to split the data fields into groups and run a PLM for each one separately.

\revised{
\subsection{Ethically Boosting Performance through Synthetic Data Augmentation}
Self-tracking data is inherently sensitive to privacy issues because they contain personal health and activity history. Therefore, training machine learning models with self-tracking data from multiple people may raise ethical issues~\cite{Saltz2019EthicsML} and thus is impractical. In contrast, our framework leverages only synthetic samples and the user's own data points to boost up model performance. Our approach demonstrates a feasibility of leveraging common sense of large language models instead of training a baseline model using data collected from a group of people. Future work remains to investigate the external validity of the synthetic samples when the framework serves real-world cases.
}

\revised{
\subsection{Warm-Starting Self-Tracking in Cold-Start Settings}
We note that the in-context learning zero-shot cases in our experiment provide \textit{pure} zero-shot experiences from the users' perspectives; with the framework embedded in a self-tracking tool, the tool is likely to yield the boosted performance even when the user inserts a natural language query for the first time. Going further, since the performance significantly increases with only one or two contextual samples (See \autoref{table:main_eval}), the user interfaces can be designed to preemptively retrieve a few samples from a new user. For example, the system may nudge the user to provide several example utterances in the initial calibration stage. Designing effective warm-starting interaction warrants further research especially from the HCI perspective.
}

\subsection{Future Application Domains}
Introducing a new NLP task for self-tracking, we propose several application domains to which our approach can be expanded further.

\paragraph{Designing User Interfaces}
Our topic-agnostic framework can be integrated to a wide range of self-tracking tools in various form factors that support natural language interaction. With speech, the framework can be employed to implement vision- and hands-free tools on smart speakers or smartwatches.
Since more than 40\% of the trials include erroneous extractions (see JGA in \autoref{table:main_eval}), proper error recovery methods (\eg, a roll-back button~[\citealt{Kim2021Data@Hand}]) should be provided to users for sustainable interaction.

\paragraph{Multi-Turn Conversation for Data Collection}
Using a tracker with a long list of data fields, it is not natural to describe all the required field values in a single utterance. Since our framework assumes that the input phrase describes a \textbf{subset} of data fields, we can expand the task as a \textbf{multi-turn conversation} scenario (\eg, \citealt{Bae2022CareCall}) where the system asks back to fill out missing information in an item. \revised{\autoref{fig:multiturn} illustrates a scenario of beer logging through a conversational agent, embedding our framework combined with a question generator. This can be also viewed as schema-guided \textit{dialog state tracking}~\cite{Rastogi2020SchemaGuided},} but the extraction of multiple-choice and long-form text fields poses challenges from the NLP perspective.

\paragraph{Schema-Free Data Collection}
We are also investigating a more radical scenario where people capture logs even without a predefined tracker schema and the system automatically generates the proper schema based on the natural language phrases.
Supporting such schema-free tracking would effectively reduce the learning curve for newcomers to self-tracking tools, especially when the trackers are customizable~\cite{Kim2017OmniTrack}.
\section{Conclusion}
In this work, we introduced a novel NLP task for data collection in self-tracking and presented an NLU framework that can effectively solve this task by augmenting PLM's prompt with synthetic samples. \revised{Drawing on the favorable outcomes from the preliminary evaluation, we discussed future research directions regarding improving the pipeline as well as designing user interfaces to effectively support self-tracking through our framework.} 
\revised{As an interdisciplinary team of both HCI and NLP researchers,} we hope this work inspires other researchers working on the growing areas of \revised{self-tracking and personal informatics, where we still need more synergistic collaboration between the NLP and HCI disciplines.}

\bibliography{main}
\bibliographystyle{acl_natbib}

\onecolumn
\appendix
\section{Exemplary Prompt for GPT-3}
\label{appendix:prompt}
\includegraphics[width=\textwidth]{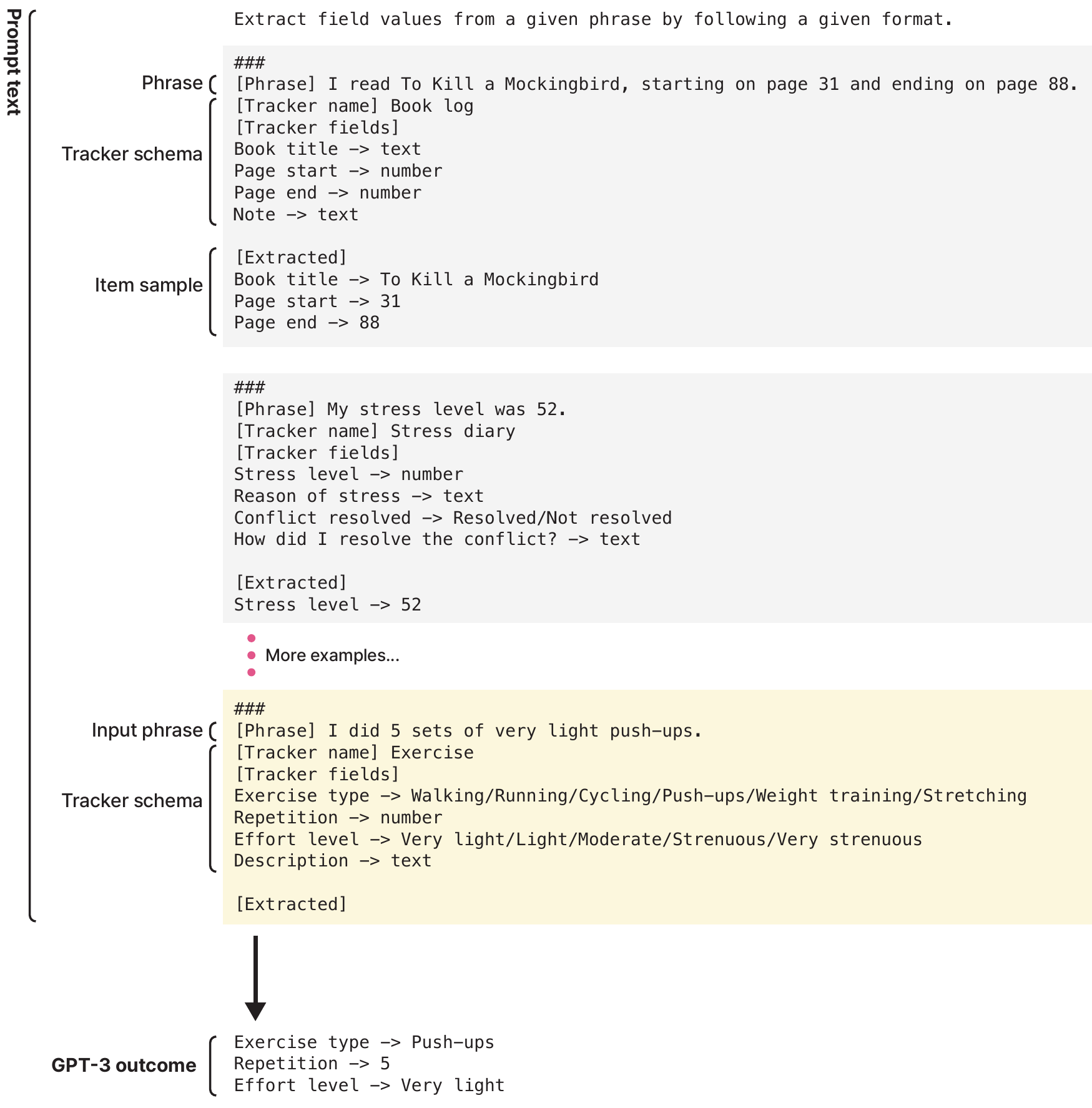}

\section{Seed Trackers}
\label{appendix:seed_trackers}
We manually composed 24 seed trackers (\ie, data schema for tracking). We first extracted 10 tracking themes from a survey of self-tracking research~\cite{Epstein2020Mapping} and an empirical study on user-defined self-trackers~\cite{Kim2017OmniTrack}. Of the four common types of trackers---\textit{timespamper}, \textit{in-situ experience logger}, \textit{daily summary}, and \textit{archive}---identified by \citealt{Kim2017OmniTrack}, we composed trackers that fall within either  \textit{in-situ experience logger} (A data entry denotes one unit of event or episode) or \textit{daily summary} (A data entry denotes a summarized reflection or information of a day). When designing schemas, we referred to prior self-tracking research prototypes or commercial apps that we have public access to the data format. \autoref{table:seed_trackers} summarizes the format of all seed trackers. Note that all trackers include one time-related field (\eg, Date, Time-point, Time-range), which we omitted in the table for brevity.
\newpage
\def\fieldcolumnwidth{0.6\textwidth}
\def\fieldarraystretch{1.2}

    \sffamily
    \small
    \centering
	\def\arraystretch{1.7}
    \setlength{\tabcolsep}{0.2em}
    \begin{longtable}{llp{\fieldcolumnwidth}l}
    
    \caption{List of the seed trackers that were used for creating synthetic item samples. The Reference column denotes the existing research or commercial apps that informed the design of the schema.} \label{table:seed_trackers} \\

\hline
\multicolumn{2}{l}{\textbf{Type/Name}} & \textbf{Data Fields} & \textbf{Reference} \\
\thicktablehline
\endfirsthead

\multicolumn{4}{l}%
{{\tablename\ \thetable{} -- continued from the previous page}} \\
\hline
\multicolumn{2}{l}{\textbf{Type/Name}} & \textbf{Data Fields} & \textbf{Reference} \\
\thicktablehline
\endhead

\hline \multicolumn{4}{r}{{$\downarrow$ Continued on the next page}} \\
\endfoot

\hline \hline
\endlastfoot


\multicolumn{4}{l}{\parbox[t]{\textwidth}{\vspace{1mm}\textbf{Exercise}\vspace{2pt}}} \\ \thicktablehline

\trackerrowcolor\lfbox[boxinsitu]{In-situ} & \textbf{Exercise} & 

\parbox[t]{\fieldcolumnwidth}{\begin{itemize*}[label={}, leftmargin=*]
\item \def\arraystretch{\fieldarraystretch}\begin{tabular}{l}\footnotesize{Exercise type} \\ \textcolor{gray}{\scriptsize{Choice-single}} \end{tabular}
\item \def\arraystretch{\fieldarraystretch}\begin{tabular}{l}\footnotesize{Repetition} \\ \textcolor{gray}{\scriptsize{Number}} \end{tabular}
\item \def\arraystretch{\fieldarraystretch}\begin{tabular}{l}\footnotesize{Intensity} \\ \textcolor{gray}{\scriptsize{Scale}} \end{tabular}
\item \def\arraystretch{\fieldarraystretch}\begin{tabular}{l}\footnotesize{Description} \\ \textcolor{gray}{\scriptsize{Text}} \end{tabular}
\end{itemize*}} & 
\scriptsize{\citealt{Kim2022MyMove}} \\

\tablelighthline

\trackerrowcolor\lfbox[boxdaily]{Daily} & \textbf{Daily exercise} &

\parbox[t]{\fieldcolumnwidth}{\begin{itemize*}[label={}, leftmargin=*, itemjoin=]
\item \def\arraystretch{\fieldarraystretch}\begin{tabular}{l}\footnotesize{Exercise done today} \\ \textcolor{gray}{\scriptsize{Choice-multiple}} \end{tabular}
\item \def\arraystretch{\fieldarraystretch}\begin{tabular}{l}\footnotesize{Overall satisfaction} \\ \textcolor{gray}{\scriptsize{Number}} \end{tabular} \\
\item \def\arraystretch{\fieldarraystretch}\begin{tabular}{l}\footnotesize{Reflections on today's exercise} \\ \textcolor{gray}{\scriptsize{Text}} \end{tabular}
\end{itemize*}} & \\

\tablelighthline

\multicolumn{4}{l}{\parbox[t]{\textwidth}{\vspace{1mm}\textbf{Sleep}\vspace{2pt}}} \\ \thicktablehline

\trackerrowcolor\lfbox[boxdaily]{Daily} & \textbf{Sleep diary} &
\parbox[t]{\fieldcolumnwidth}{\begin{itemize*}[label={}, leftmargin=*]
\item \def\arraystretch{\fieldarraystretch}\begin{tabular}{l}\footnotesize{Sleep quality} \\ \textcolor{gray}{\scriptsize{Scale}} \end{tabular}
\item \def\arraystretch{\fieldarraystretch}\begin{tabular}{l}\footnotesize{Memo} \\ \textcolor{gray}{\scriptsize{Text}} \end{tabular}
\end{itemize*}} & \\ 

\tablelighthline

\multicolumn{4}{l}{\parbox[t]{\textwidth}{\vspace{1mm}\textbf{Medication}\vspace{2pt}}} \\ \thicktablehline

\trackerrowcolor\lfbox[boxinsitu]{In-situ} & \textbf{Pill intake} &
\parbox[t]{\fieldcolumnwidth}{\begin{itemize*}[label={}, leftmargin=*]
\item \def\arraystretch{\fieldarraystretch}\begin{tabular}{l}\footnotesize{Medication} \\ \textcolor{gray}{\scriptsize{Choice-single}} \end{tabular}
\item \def\arraystretch{\fieldarraystretch}\begin{tabular}{l}\footnotesize{Number of pills} \\ \textcolor{gray}{\scriptsize{Number}} \end{tabular}
\item \def\arraystretch{\fieldarraystretch}\begin{tabular}{l}\footnotesize{Reason of taking} \\ \textcolor{gray}{\scriptsize{Text}} \end{tabular}
\end{itemize*}} & \\ \tablelighthline

\multicolumn{4}{l}{\parbox[t]{\textwidth}{\vspace{1mm}\textbf{Diabetes}\vspace{2pt}}} \\ \thicktablehline

\trackerrowcolor\lfbox[boxinsitu]{In-situ} & \textbf{Insulin shots} &
\parbox[t]{\fieldcolumnwidth}{\begin{itemize*}[label={}, leftmargin=*]
\item \def\arraystretch{\fieldarraystretch}\begin{tabular}{l}\footnotesize{Type} \\ \textcolor{gray}{\scriptsize{Choice-single}} \end{tabular}
\item \def\arraystretch{\fieldarraystretch}\begin{tabular}{l}\footnotesize{Units} \\ \textcolor{gray}{\scriptsize{Number}} \end{tabular}
\end{itemize*}} & \scriptsize{\citealt{MyNetDiary}} \\ \tablelighthline

\trackerrowcolor\lfbox[boxinsitu]{In-situ} & \textbf{Blood sugar} &
\parbox[t]{\fieldcolumnwidth}{\begin{itemize*}[label={}, leftmargin=*]
\item \def\arraystretch{\fieldarraystretch}\begin{tabular}{l}\footnotesize{Glucose level} \\ \textcolor{gray}{\scriptsize{Number}} \end{tabular}
\item \def\arraystretch{\fieldarraystretch}\begin{tabular}{l}\footnotesize{Measurement Timing} \\ \textcolor{gray}{\scriptsize{Choice-single}} \end{tabular}
\end{itemize*}} & \scriptsize{\citealt{MyNetDiary}} \\ \tablelighthline

\multicolumn{4}{l}{\parbox[t]{\textwidth}{\vspace{1mm}\textbf{Food}\vspace{2pt}}} \\ \thicktablehline

\trackerrowcolor\lfbox[boxinsitu]{In-situ} & \textbf{Meal log} &
\parbox[t]{\fieldcolumnwidth}{\begin{itemize*}[label={}, leftmargin=*]
\item \def\arraystretch{\fieldarraystretch}\begin{tabular}{l}\footnotesize{Meal type} \\ \textcolor{gray}{\scriptsize{Choice-single}} \end{tabular}
\item \def\arraystretch{\fieldarraystretch}\begin{tabular}{l}\footnotesize{Menu} \\ \textcolor{gray}{\scriptsize{Text}} \end{tabular}
\item \def\arraystretch{\fieldarraystretch}\begin{tabular}{l}\footnotesize{Why I ate this food} \\ \textcolor{gray}{\scriptsize{Text}} \end{tabular}
\item \def\arraystretch{\fieldarraystretch}\begin{tabular}{l}\footnotesize{Healthy level} \\ \textcolor{gray}{\scriptsize{Scale}} \end{tabular}
\end{itemize*}} & \scriptsize{\citealt{Luo2021FoodScrap}} \\ \tablelighthline

\trackerrowcolor\lfbox[boxinsitu]{In-situ} & \textbf{Visited restaurant} &
\parbox[t]{\fieldcolumnwidth}{\begin{itemize*}[label={}, leftmargin=*]
\item \def\arraystretch{\fieldarraystretch}\begin{tabular}{l}\footnotesize{Restaurant name} \\ \textcolor{gray}{\scriptsize{Text}} \end{tabular}
\item \def\arraystretch{\fieldarraystretch}\begin{tabular}{l}\footnotesize{Type of cuisine} \\ \textcolor{gray}{\scriptsize{Choice-single}} \end{tabular}
\item \def\arraystretch{\fieldarraystretch}\begin{tabular}{l}\footnotesize{Menus I tried} \\ \textcolor{gray}{\scriptsize{Text}} \end{tabular}
\item \def\arraystretch{\fieldarraystretch}\begin{tabular}{l}\footnotesize{Rating for taste} \\ \textcolor{gray}{\scriptsize{Number}} \end{tabular}
\item \def\arraystretch{\fieldarraystretch}\begin{tabular}{l}\footnotesize{Rating for location} \\ \textcolor{gray}{\scriptsize{Number}} \end{tabular}
\item \def\arraystretch{\fieldarraystretch}\begin{tabular}{l}\footnotesize{Rating for hygiene} \\ \textcolor{gray}{\scriptsize{Number}} \end{tabular}
\item \def\arraystretch{\fieldarraystretch}\begin{tabular}{l}\footnotesize{Rating for staff} \\ \textcolor{gray}{\scriptsize{Number}} \end{tabular}
\item \def\arraystretch{\fieldarraystretch}\begin{tabular}{l}\footnotesize{Review} \\ \textcolor{gray}{\scriptsize{Text}} \end{tabular}
\end{itemize*}}& \\ \tablelighthline

\trackerrowcolor\lfbox[boxdaily]{Daily} & \textbf{Daily eating} &
\parbox[t]{\fieldcolumnwidth}{\begin{itemize*}[label={}, leftmargin=*]
\item \def\arraystretch{\fieldarraystretch}\begin{tabular}{l}\footnotesize{Types of meals had} \\ \textcolor{gray}{\scriptsize{Choice-multiple}} \end{tabular}
\item \def\arraystretch{\fieldarraystretch}\begin{tabular}{l}\footnotesize{Relfection on today's eating} \\ \textcolor{gray}{\scriptsize{Text}} \end{tabular}
\end{itemize*}} & \\ \tablelighthline

\multicolumn{4}{l}{\parbox[t]{\textwidth}{\vspace{1mm}\textbf{Beverage}\vspace{2pt}}} \\ \thicktablehline

\trackerrowcolor\lfbox[boxinsitu]{In-situ} & \textbf{Beverage log} &
\parbox[t]{\fieldcolumnwidth}{\begin{itemize*}[label={}, leftmargin=*]
\item \def\arraystretch{\fieldarraystretch}\begin{tabular}{l}\footnotesize{Category} \\ \textcolor{gray}{\scriptsize{Choice-single}} \end{tabular}
\item \def\arraystretch{\fieldarraystretch}\begin{tabular}{l}\footnotesize{Name} \\ \textcolor{gray}{\scriptsize{Text}} \end{tabular}
\item \def\arraystretch{\fieldarraystretch}\begin{tabular}{l}\footnotesize{Temperature} \\ \textcolor{gray}{\scriptsize{Choice-single}} \end{tabular}
\item \def\arraystretch{\fieldarraystretch}\begin{tabular}{l}\footnotesize{Cups} \\ \textcolor{gray}{\scriptsize{Number}} \end{tabular}
\item \def\arraystretch{\fieldarraystretch}\begin{tabular}{l}\footnotesize{Location} \\ \textcolor{gray}{\scriptsize{Choice-single}} \end{tabular}
\end{itemize*}} & \\ \tablelighthline

\trackerrowcolor\lfbox[boxinsitu]{In-situ} & \textbf{Beer log} &
\parbox[t]{\fieldcolumnwidth}{\begin{itemize*}[label={}, leftmargin=*]
\item \def\arraystretch{\fieldarraystretch}\begin{tabular}{l}\footnotesize{Name} \\ \textcolor{gray}{\scriptsize{Text}} \end{tabular}
\item \def\arraystretch{\fieldarraystretch}\begin{tabular}{l}\footnotesize{Beer category} \\ \textcolor{gray}{\scriptsize{Choice-single}} \end{tabular}
\item \def\arraystretch{\fieldarraystretch}\begin{tabular}{l}\footnotesize{Score} \\ \textcolor{gray}{\scriptsize{Number}} \end{tabular}
\item \def\arraystretch{\fieldarraystretch}\begin{tabular}{l}\footnotesize{Review} \\ \textcolor{gray}{\scriptsize{Text}} \end{tabular}
\end{itemize*}} & \scriptsize{\citealt{Kim2017OmniTrack}} \\ \tablelighthline

\trackerrowcolor\lfbox[boxdaily]{Daily} & \textbf{Daily coffee} &
\parbox[t]{\fieldcolumnwidth}{\begin{itemize*}[label={}, leftmargin=*]
\item \def\arraystretch{\fieldarraystretch}\begin{tabular}{l}\footnotesize{Number of cups had today} \\ \textcolor{gray}{\scriptsize{Number}} \end{tabular}
\item \def\arraystretch{\fieldarraystretch}\begin{tabular}{l}\footnotesize{Why I had that amount of coffee} \\ \textcolor{gray}{\scriptsize{Text}} \end{tabular}
\end{itemize*}} & \scriptsize{\citealt{Luo2021FoodScrap}} \\ \tablelighthline

\multicolumn{4}{l}{\parbox[t]{\textwidth}{\vspace{1mm}\textbf{Mood}\vspace{2pt}}} \\ \thicktablehline

\trackerrowcolor\lfbox[boxinsitu]{In-situ} & \textbf{Mood episodes} & \parbox[t]{\fieldcolumnwidth}{\begin{itemize*}[label={}, leftmargin=*]
\item \def\arraystretch{\fieldarraystretch}\begin{tabular}{l}\footnotesize{Types of mood} \\ \textcolor{gray}{\scriptsize{Choice-multiple}} \end{tabular}
\item \def\arraystretch{\fieldarraystretch}\begin{tabular}{l}\footnotesize{Intensity of mood} \\ \textcolor{gray}{\scriptsize{Number}} \end{tabular}
\item \def\arraystretch{\fieldarraystretch}\begin{tabular}{l}\footnotesize{Reason of mood} \\ \textcolor{gray}{\scriptsize{Text}} \end{tabular}
\end{itemize*}} & \\ \tablelighthline

\trackerrowcolor\lfbox[boxinsitu]{In-situ} & \textbf{Stress diary} &
\parbox[t]{0.45\textwidth}{\begin{itemize*}[label={}, leftmargin=*]
\item \def\arraystretch{\fieldarraystretch}\begin{tabular}{l}\footnotesize{Stress level} \\ \textcolor{gray}{\scriptsize{Number}} \end{tabular}
\item \def\arraystretch{\fieldarraystretch}\begin{tabular}{l}\footnotesize{Reason of stress} \\ \textcolor{gray}{\scriptsize{Text}} \end{tabular}
\item \def\arraystretch{\fieldarraystretch}\begin{tabular}{l}\footnotesize{Conflict resolved} \\ \textcolor{gray}{\scriptsize{Choice-single}} \end{tabular}
\item \def\arraystretch{\fieldarraystretch}\begin{tabular}{l}\footnotesize{How did I resolve the conflict?} \\ \textcolor{gray}{\scriptsize{Text}} \end{tabular}
\end{itemize*}} & \scriptsize{\citealt{Dietz2019StressAnnotation}} \\ \tablelighthline

\trackerrowcolor\lfbox[boxdaily]{Daily} & \textbf{Daily diary} &
\parbox[t]{0.56\textwidth}{\begin{itemize*}[label={}, leftmargin=*]
\item \def\arraystretch{\fieldarraystretch}\begin{tabular}{l}\footnotesize{Weather} \\ \textcolor{gray}{\scriptsize{Choice-single}} \end{tabular}
\item \def\arraystretch{\fieldarraystretch}\begin{tabular}{l}\footnotesize{Stress level} \\ \textcolor{gray}{\scriptsize{Number}} \end{tabular}
\item \def\arraystretch{\fieldarraystretch}\begin{tabular}{l}\footnotesize{Overall productivity} \\ \textcolor{gray}{\scriptsize{Scale}} \end{tabular}
\item \def\arraystretch{\fieldarraystretch}\begin{tabular}{l}\footnotesize{Atmosphere} \\ \textcolor{gray}{\scriptsize{Choice-single}} \end{tabular}
\item \def\arraystretch{\fieldarraystretch}\begin{tabular}{l}\footnotesize{Major types of mood} \\ \textcolor{gray}{\scriptsize{Choice-multiple}} \end{tabular}
\item \def\arraystretch{\fieldarraystretch}\begin{tabular}{l}\footnotesize{Whom I met today} \\ \textcolor{gray}{\scriptsize{Choice-multiple}} \end{tabular}
\item \def\arraystretch{\fieldarraystretch}\begin{tabular}{l}\footnotesize{Reflection on today} \\ \textcolor{gray}{\scriptsize{Text}} \end{tabular}
\end{itemize*}} & \\ \tablelighthline

\multicolumn{4}{l}{\parbox[t]{\textwidth}{\vspace{1mm}\textbf{Book}\vspace{2pt}}} \\ \thicktablehline

\trackerrowcolor\lfbox[boxinsitu]{In-situ} & \textbf{Book log} &
\parbox[t]{\fieldcolumnwidth}{\begin{itemize*}[label={}, leftmargin=*]
\item \def\arraystretch{\fieldarraystretch}\begin{tabular}{l}\footnotesize{Book title} \\ \textcolor{gray}{\scriptsize{Text}} \end{tabular}
\item \def\arraystretch{\fieldarraystretch}\begin{tabular}{l}\footnotesize{Page start} \\ \textcolor{gray}{\scriptsize{Number}} \end{tabular}
\item \def\arraystretch{\fieldarraystretch}\begin{tabular}{l}\footnotesize{Page end} \\ \textcolor{gray}{\scriptsize{Number}} \end{tabular}
\item \def\arraystretch{\fieldarraystretch}\begin{tabular}{l}\footnotesize{Note} \\ \textcolor{gray}{\scriptsize{Text}} \end{tabular}
\end{itemize*}} & \\ \tablelighthline

\multicolumn{4}{l}{\parbox[t]{\textwidth}{\vspace{1mm}\textbf{Study}\vspace{2pt}}} \\ \thicktablehline

\trackerrowcolor\lfbox[boxinsitu]{In-situ} & \textbf{Study log} & \parbox[t]{\fieldcolumnwidth}{\begin{itemize*}[label={}, leftmargin=*]
\item \def\arraystretch{\fieldarraystretch}\begin{tabular}{l}\footnotesize{Study subject} \\ \textcolor{gray}{\scriptsize{Choice-single}} \end{tabular}
\item \def\arraystretch{\fieldarraystretch}\begin{tabular}{l}\footnotesize{Accomplishment} \\ \textcolor{gray}{\scriptsize{Number}} \end{tabular}
\item \def\arraystretch{\fieldarraystretch}\begin{tabular}{l}\footnotesize{Study content} \\ \textcolor{gray}{\scriptsize{Text}} \end{tabular}
\end{itemize*}} & \\ \tablelighthline

\trackerrowcolor\lfbox[boxdaily]{Daily} & \textbf{Study diary} & 
\parbox[t]{\fieldcolumnwidth}{\begin{itemize*}[label={}, leftmargin=*]
\item \def\arraystretch{\fieldarraystretch}\begin{tabular}{l}\footnotesize{Study subjects} \\ \textcolor{gray}{\scriptsize{Choice-multiple}} \end{tabular}
\item \def\arraystretch{\fieldarraystretch}\begin{tabular}{l}\footnotesize{Overall satisfaction} \\ \textcolor{gray}{\scriptsize{Number}} \end{tabular}
\item \def\arraystretch{\fieldarraystretch}\begin{tabular}{l}\footnotesize{Reflection on today's study} \\ \textcolor{gray}{\scriptsize{Text}} \end{tabular}
\end{itemize*}} & \\ \tablelighthline

\multicolumn{4}{l}{\parbox[t]{\textwidth}{\vspace{1mm}\textbf{Productivity}\vspace{2pt}}} \\ \thicktablehline

\trackerrowcolor\lfbox[boxinsitu]{In-situ} & \textbf{Tasks} &
\parbox[t]{\fieldcolumnwidth}{\begin{itemize*}[label={}, leftmargin=*]
\item \def\arraystretch{\fieldarraystretch}\begin{tabular}{l}\footnotesize{Task} \\ \textcolor{gray}{\scriptsize{Choice-multiple}} \end{tabular}
\item \def\arraystretch{\fieldarraystretch}\begin{tabular}{l}\footnotesize{Productivity} \\ \textcolor{gray}{\scriptsize{Scale}} \end{tabular}
\item \def\arraystretch{\fieldarraystretch}\begin{tabular}{l}\footnotesize{Rationale for productivity} \\ \textcolor{gray}{\scriptsize{Text}} \end{tabular}
\end{itemize*}} & \scriptsize{\citealt{Kim2019UnderstandingProductivity}} \\ \tablelighthline

\trackerrowcolor\lfbox[boxinsitu]{In-situ} & \textbf{Breaks} &
\parbox[t]{\fieldcolumnwidth}{\begin{itemize*}[label={}, leftmargin=*]
\item \def\arraystretch{\fieldarraystretch}\begin{tabular}{l}\footnotesize{What I did during the break} \\ \textcolor{gray}{\scriptsize{Choice-single}} \end{tabular}
\item \def\arraystretch{\fieldarraystretch}\begin{tabular}{l}\footnotesize{Reason for break} \\ \textcolor{gray}{\scriptsize{Text}} \end{tabular}
\end{itemize*}} & \scriptsize{\citealt{Epstein2016Taking5}} \\ \tablelighthline

\trackerrowcolor\lfbox[boxdaily]{Daily} & \textbf{Work diary} &
\parbox[t]{\fieldcolumnwidth}{\begin{itemize*}[label={}, leftmargin=*]
\item \def\arraystretch{\fieldarraystretch}\begin{tabular}{l}\footnotesize{Major tasks} \\ \textcolor{gray}{\scriptsize{Choice-multiple}} \end{tabular}
\item \def\arraystretch{\fieldarraystretch}\begin{tabular}{l}\footnotesize{Overall productivity} \\ \textcolor{gray}{\scriptsize{Number}} \end{tabular}
\item \def\arraystretch{\fieldarraystretch}\begin{tabular}{l}\footnotesize{Reflections on today's work} \\ \textcolor{gray}{\scriptsize{Text}} \end{tabular}
\end{itemize*}} & \\ \tablelighthline

\multicolumn{4}{l}{\parbox[t]{\textwidth}{\vspace{1mm}\textbf{Social}\vspace{2pt}}} \\ \thicktablehline

\trackerrowcolor\lfbox[boxinsitu]{In-situ} & \textbf{People} &
\parbox[t]{0.4\textwidth}{\begin{itemize*}[label={}, leftmargin=*]
\item \def\arraystretch{\fieldarraystretch}\begin{tabular}{l}\footnotesize{Who I met} \\ \textcolor{gray}{\scriptsize{Text}} \end{tabular}
\item \def\arraystretch{\fieldarraystretch}\begin{tabular}{l}\footnotesize{Purpose} \\ \textcolor{gray}{\scriptsize{Choice-single}} \end{tabular}
\item \def\arraystretch{\fieldarraystretch}\begin{tabular}{l}\footnotesize{What I did in detail} \\ \textcolor{gray}{\scriptsize{Text}} \end{tabular}
\item \def\arraystretch{\fieldarraystretch}\begin{tabular}{l}\footnotesize{Reflections on the interaction} \\ \textcolor{gray}{\scriptsize{Text}} \end{tabular}
\end{itemize*}} & \\ \tablelighthline

\multicolumn{4}{l}{\parbox[t]{\textwidth}{\vspace{1mm}\textbf{Smoking}\vspace{2pt}}} \\ \thicktablehline

\trackerrowcolor\lfbox[boxinsitu]{In-situ} & \textbf{Smoking log} &
\parbox[t]{\fieldcolumnwidth}{\begin{itemize*}[label={}, leftmargin=*]
\item \def\arraystretch{\fieldarraystretch}\begin{tabular}{l}\footnotesize{Amount} \\ \textcolor{gray}{\scriptsize{Number}} \end{tabular}
\item \def\arraystretch{\fieldarraystretch}\begin{tabular}{l}\footnotesize{Tobacco name} \\ \textcolor{gray}{\scriptsize{Choice-single}} \end{tabular}
\item \def\arraystretch{\fieldarraystretch}\begin{tabular}{l}\footnotesize{Smoking context} \\ \textcolor{gray}{\scriptsize{Choice-single}} \end{tabular}
\item \def\arraystretch{\fieldarraystretch}\begin{tabular}{l}\footnotesize{Smoked with others} \\ \textcolor{gray}{\scriptsize{Choice-single}} \end{tabular}
\end{itemize*}} & \\ \tablelighthline

\multicolumn{4}{l}{\parbox[t]{\textwidth}{\vspace{1mm}\textbf{Female}\vspace{2pt}}} \\ \thicktablehline

\trackerrowcolor\lfbox[boxdaily]{Daily} & \textbf{Daily period} & \parbox[t]{\fieldcolumnwidth}{\begin{itemize*}[label={}, leftmargin=*]
\item \def\arraystretch{\fieldarraystretch}\begin{tabular}{l}\footnotesize{Bleeding} \\ \textcolor{gray}{\scriptsize{Choice-single}} \end{tabular}
\item \def\arraystretch{\fieldarraystretch}\begin{tabular}{l}\footnotesize{Pain} \\ \textcolor{gray}{\scriptsize{Choice-multiple}} \end{tabular}
\item \def\arraystretch{\fieldarraystretch}\begin{tabular}{l}\footnotesize{Emotions} \\ \textcolor{gray}{\scriptsize{Choice-multiple}} \end{tabular}
\item \def\arraystretch{\fieldarraystretch}\begin{tabular}{l}\footnotesize{Sleep} \\ \textcolor{gray}{\scriptsize{Choice-single}} \end{tabular}
\item \def\arraystretch{\fieldarraystretch}\begin{tabular}{l}\footnotesize{Sex} \\ \textcolor{gray}{\scriptsize{Choice-multiple}} \end{tabular}
\item \def\arraystretch{\fieldarraystretch}\begin{tabular}{l}\footnotesize{Energy} \\ \textcolor{gray}{\scriptsize{Choice-single}} \end{tabular}
\item \def\arraystretch{\fieldarraystretch}\begin{tabular}{l}\footnotesize{Social} \\ \textcolor{gray}{\scriptsize{Choice-single}} \end{tabular}
\item \def\arraystretch{\fieldarraystretch}\begin{tabular}{l}\footnotesize{Reflection on today's menstruation} \\ \textcolor{gray}{\scriptsize{Text}} \end{tabular}
\end{itemize*}} & \scriptsize{\citealt{ClueApp}} \\

\end{longtable}

\end{document}